\pdfoutput=1

\documentclass[11pt]{article}

\usepackage{EMNLP2022}

\usepackage{times}
\usepackage{latexsym}

\usepackage[T1]{fontenc}

\usepackage[utf8]{inputenc}

\usepackage{microtype}

\usepackage{inconsolata}

\usepackage{subfigure}
\usepackage{xfrac}
\usepackage{colortbl}
\usepackage{booktabs}  
\usepackage{amsfonts}  
\usepackage{nicefrac}
\usepackage{latexsym}
\usepackage{multirow}
\usepackage{amsmath}
\usepackage{tabularx}
\usepackage{makecell}
\usepackage{xcolor}

\definecolor{gred}{RGB}{219,68,55}
\definecolor{gblue}{RGB}{66,133,244}
\definecolor{gyellow}{RGB}{244,180,0}
\definecolor{ggreen}{RGB}{85,157,88}
\definecolor{ggrey}{RGB}{115,115,115}
\definecolor{na}{gray}{0.9}

\newcommand{\colorR}[1]{\textcolor{gred}{#1}}
\newcommand{\colorG}[1]{\textcolor{ggreen}{#1}}
\newcommand{\colorB}[1]{\textcolor{gblue}{#1}}
\newcommand{\colorY}[1]{\textcolor{gyellow}{#1}}

\newcommand*{\affmark}[1][*]{\textsuperscript{#1}}
\newcommand*{\email}[1]{\texttt{#1}}

\usepackage{xparse}
\NewDocumentCommand{\heng}
{ mO{} }{\textcolor{red}{\textsuperscript{\textit{Heng}}\textsf{\textbf{\small[#1]}}}}

\title{Towards a Unified Multi-Dimensional Evaluator for Text Generation}

\author{
Ming Zhong\affmark[$\S$]
\bf \quad Yang Liu\affmark[$\dagger$]
\quad Da Yin\affmark[$\clubsuit$]
\quad Yuning Mao\affmark[$\S$]
\quad Yizhu Jiao\affmark[$\S$]\\
\bf \quad Pengfei Liu\affmark[$\ddagger$] 
\bf \quad Chenguang Zhu\affmark[$\dagger$]
\bf \quad Heng Ji\affmark[$\S$]
\bf \quad Jiawei Han\affmark[$\S$]
\\
{\affmark[$\S$]University of Illinois at Urbana-Champaign} \quad
{\affmark[$\dagger$]Microsoft Cognitive Services Research}\\
{\affmark[$\clubsuit$]University of California, Los Angeles} \quad {\affmark[$\ddagger$]Carnegie Mellon University} \\
\email{\{mingz5, yuningm2, yizhuj2, hengji, hanj\}@illinois.edu}\\
\email{\{yaliu10, chezhu\}}@microsoft.com \quad da.yin@cs.ucla.edu
\quad pliu3@cs.cmu.edu}

\begin{document}
\maketitle
\begin{abstract}

Multi-dimensional evaluation is the dominant paradigm for human evaluation in Natural Language Generation (NLG), i.e., evaluating the generated text from multiple explainable dimensions, such as coherence and fluency.
However, automatic evaluation in NLG is still dominated by similarity-based metrics, and we lack a reliable framework for a more comprehensive evaluation of advanced models.
In this paper, we propose a unified multi-dimensional evaluator \textsc{UniEval} for NLG.
We re-frame NLG evaluation as a Boolean Question Answering (QA) task, and by guiding the model with different questions, we can use one evaluator to evaluate from multiple dimensions.
Furthermore, thanks to the unified Boolean QA format, we are able to introduce an intermediate learning phase that enables \textsc{UniEval} to incorporate external knowledge from multiple related tasks and gain further improvement.
Experiments on three typical NLG tasks show that \textsc{UniEval} correlates substantially better with human judgments than existing metrics.
Specifically, compared to the top-performing unified evaluators, \textsc{UniEval} achieves a 23\% higher correlation on text summarization, and over 43\% on dialogue response generation.
Also, \textsc{UniEval} demonstrates a strong zero-shot learning ability for unseen evaluation dimensions and tasks.
Source code, data and all  pre-trained evaluators are available on our GitHub repository\footnote{\url{https://github.com/maszhongming/UniEval}}.
\end{abstract}

\section{Introduction}


The rapid development of Natural Language Generation (NLG) tasks with the support of pre-trained language models~\cite{raffel2020exploring,brown2020language,DBLP:conf/acl/LewisLGGMLSZ20} calls for a higher quality evaluation of generated texts.
However, the evaluation process is still dominated by traditional similarity-based metrics~\cite{kasai2021bidimensional}, exemplified by ROUGE~\cite{lin2004rouge} and BLEU~\cite{papineni2002bleu} that compute n-gram overlap between the model output and the reference text.
These metrics are potentially misleading as NLG models have advanced to the point where discrepancies between them are unlikely to be detected based on surface-level features~\cite{gehrmann2022repairing}.
Although using pre-trained models to obtain embedding-based similarity may alleviate this issue~\cite{zhang2019bertscore}, these metrics still naturally lead to the question: \textit{does similarity to reference text indicate the overall quality of model output?}
\citet{belz2008intrinsic} referred to this similarity as ``human-likeness'' and pointed out that the ability to output human-like text may be completely unrelated to the final performance on generation tasks.

Realizing that creating a one-size-fits-all score is infeasible, subsequent research has focused on a more comprehensive multi-dimensional evaluation for NLG tasks.
It aims to evaluate the model output from multiple explainable dimensions and has been the dominant paradigm in human evaluation~\cite{fabbri2021summeval}.
For example, text summarization typically uses four dimensions for evaluation: \texttt{coherence}, \texttt{consistency}, \texttt{fluency}, and \texttt{relevance} (see Table~\ref{tab:case}).
One way to achieve this fine-grained evaluation is to develop multiple evaluators dedicated to every single dimension~\cite{dziri2019evaluating, kryscinski2020evaluating}.
However, it requires extensive effort to individually select and train an evaluator for each dimension when conducting multi-dimensional evaluations.
On the other hand, several studies worked on building a unified evaluator, i.e., a single model that can produce multiple metrics (e.g., precision and recall) for the generated text~\cite{yuan2021bartscore}.
Nevertheless, their evaluation scores cannot be directly aligned with the dimensions designed in human evaluation (e.g., \texttt{consistency} and \texttt{coherence}).

\renewcommand\arraystretch{1.0}
\begin{table*}
\centering \footnotesize
\label{tab:pattern}
\begin{tabular}{p{15cm}}
\toprule
\textbf{Generated Summary}: Harry Kane is nominated for both the PFA player and young player of the season. The Spurs striker has been released from the awards ceremony on Sunday. The Tottenham striker features in a new animation. \\
\textbf{Reference Summary}: Harry Kane has been in superb form for Tottenham this season. The 21-year-old has scored 30 goals in all competitions for Spurs. Kane also made his England debut and scored within two minutes. \\
\textbf{Document}: Harry Kane's celebrations this season have always shown him to be an animated young man … \\
\midrule
\textbf{Similarity-based Evaluators} \\
ROUGE-1: 0.44 \qquad \qquad ROUGE-2: 0.25 \qquad \qquad ROUGE-L: 0.42 \qquad \qquad BERTScore: 0.24 \\
\midrule
\textbf{Single-dimensional Evaluators} (predicted by two different evaluators~\cite{deng2021compression}) \\
Consistency: 0.87 \qquad \quad \, Relevance: 0.74 \\
\midrule
\textbf{Unified Evaluator} (predicted by BARTScore, and the scoring range is negative infinity to 0) \\
Precision: -5.45 \hspace{4.0em} Recall: -4.93 \hspace{5.1em} $\mathrm{F}_1$: -5.19 \\
\midrule
\textbf{Multi-dimensional Evaluator} (predicted by our evaluator \textsc{UniEval}) \\
Coherence: 0.04 \hspace{3.8em} Consistency: 0.41 \hspace{3.1em} Fluency: 0.92 \hspace{4.8em} Relevance: 0.28 \\
\midrule
\textbf{Human Evaluation} (annotated by experts, and we map the scoring range to 0-1) \\
Coherence: 0.00 \hspace{3.8em} Consistency: 0.25 \hspace{3.1em} Fluency: 1.00 \hspace{4.8em} Relevance: 0.33 \\
\bottomrule
\end{tabular}
\caption{An example of evaluating the text summarization task. All metrics except BARTScore are scored in the range of 0 to 1, with higher scores indicating better quality.
Our proposed \textsc{UniEval} is consistent with human evaluation, using multiple dimensions: \texttt{Coherence}, \texttt{Consistency}, \texttt{Fluency}, and \texttt{Relevance} to evaluate the generated text. The scores predicted by \textsc{UniEval} are closer to human judgements.}
\label{tab:case}
\end{table*}

In this paper, we propose a unified multi-dimensional evaluator \textsc{UniEval} for text generation tasks.
\textsc{UniEval} unifies all evaluation dimensions into a Boolean Question Answering (QA) problem~\cite{clark2019boolq}, thus enabling the evaluation of the generated text from different perspectives using only a single model.
For instance, \textsc{UniEval} can evaluate \texttt{coherence} in summarization by inputting a specific question, such as \textit{``Is this a coherent summary to the document?''}.
Moreover, thanks to the unified Boolean QA format, we are able to perform an intermediate training stage on four types of tasks related to NLG evaluation.
This can be crucial for evaluation quality, since we lack large-scale human scores of model outputs to train an evaluator, a unified format that encompasses diverse existing tasks (namely, intermediate tasks) can substantially help \textsc{UniEval} incorporate external knowledge related to NLG evaluations.

Specifically, a unified framework can bring the following benefits:

\noindent 1) \textbf{Ease of use}. One model is sufficient, without the effort of picking multiple appropriate single-dimensional evaluators for all the dimensions.

\noindent 2) \textbf{Internal complementarity}. Different dimensions in the same NLG task can be closely related to each other, so it is useful to perform joint training for these dimensions to share knowledge.

\noindent 3) \textbf{External knowledge incorporation}. 
The unified Boolean QA format makes it possible to enhance the pre-trained language model by multi-task learning on diverse and relevant intermediate tasks before being trained on evaluation tasks.

\noindent 4) \textbf{Extensibility and transferability}. 
A unified evaluator can achieve better extensibility and transferability with continual learning~\cite{parisi2019continual} or prompting~\cite{liu2021pre, chen2022adaprompt}, as it can accommodate more evaluation dimensions by modifying the input question.

Experimentally, \textsc{UniEval} surpasses advanced evaluators by a large margin when evaluating three typical NLG tasks.
Concretely, compared to the best unified evaluators~\cite{yuan2021bartscore, mehri2020usr}, \textsc{UniEval} improves the correlation with human judgment by 23\% on text summarization, and the improvement exceeds 43\% on dialogue response generation.
Ablation studies verify the effectiveness of
our intermediate tasks.
We also conduct transfer experiments and show that \textsc{UniEval} achieves better performance compared with strong baseline metrics on unseen dimensions and NLG tasks in a zero-shot setting.

\section{Related Work}

\paragraph{Similarity-based Metrics}
Similarity-based metrics refer to the scores for evaluating the NLG models by measuring the similarity between a generated text and a reference text.
They can be divided into lexical overlap-based~\cite{papineni2002bleu, lin2004rouge, banerjee2005meteor} as well as contextualized embedding-based ~\cite{zhang2019bertscore, zhao2019moverscore, clark2019sentence} evaluators.
Although more than 60\% of recent NLG papers solely use ROUGE or BLEU as the evaluation metric~\cite{kasai2021bidimensional}, they fail to measure content quality~\cite{reiter2009investigation} and syntactic correctness~\cite{stent2005evaluating}, and are thus insufficient to portray the reliability of NLG systems.

\paragraph{Single-dimensional Evaluator}
To conduct more fine-grained evaluations for NLG, recent studies develop evaluators for a specific dimension, such as \texttt{consistency} in summarization~\cite{kryscinski2020evaluating, wang2020asking, cao2020factual, durmus2020feqa} and \texttt{coherence} in dialogue response generation~\cite{dziri2019evaluating,huang2020grade, ye2021towards}.
These evaluators can help us better understand the characteristics of advanced NLG models from different perspectives.
However, considering that most dimensions currently have no corresponding standard evaluators, solely using multiple single-dimensional evaluators to perform multi-dimensional evaluation is hard to achieve.

\paragraph{Unified Evaluator}
Several recent evaluators can predict multiple numbers for evaluating text by using different input and output contents~\cite{yuan2021bartscore}, multiple model variants~\cite{mehri2020usr}, or different formulas~\cite{scialom2021questeval}, and we refer to them as unified evaluators.
These evaluation scores usually have no corresponding explanations or are simply categorized as precision, recall, and $\mathrm{F}_1$, which poses difficulties in how to use them.
Therefore, we propose a unified multi-dimensional evaluator in this paper, which attempts to align the evaluation scores with different dimensions in human evaluation.

\section{Method}

\begin{figure*}[t]
    \centering
    \includegraphics[width=1.0\linewidth]{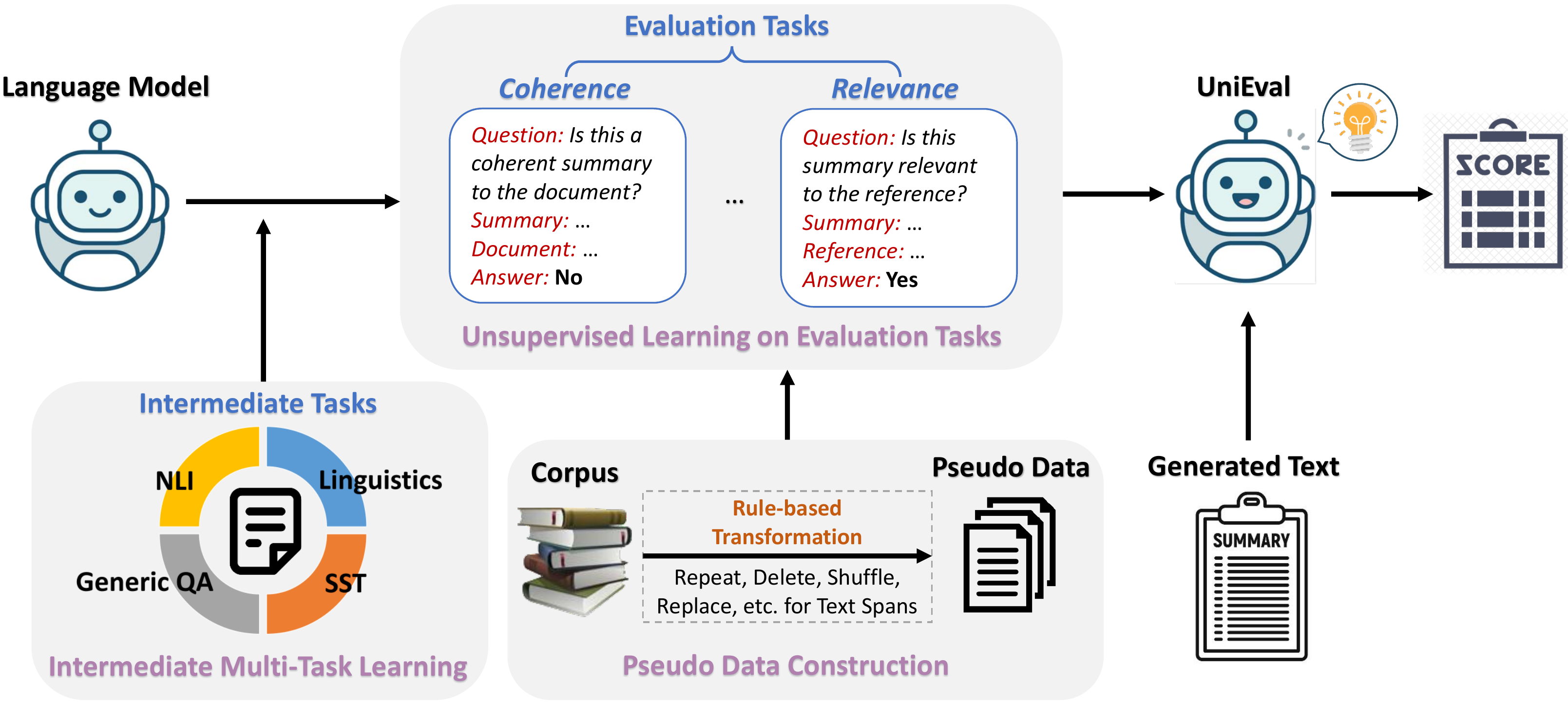}
    \caption{The overall framework of \textsc{UniEval}. We convert all tasks into a Boolean QA problem and utilize the model to answer with ``Yes” or ``No''. This unified QA format allows the model to incorporate external knowledge from multiple related tasks, i.e., intermediate multi-task learning. Then we construct pseudo data for each dimension and train them sequentially to obtain \textsc{UniEval}.}
    \label{fig:model}
\end{figure*}

In this section, we first introduce how to formulate multi-dimensional evaluation as a unified Boolean QA problem, and then describe in detail the training paradigm for \textsc{UniEval}.

\subsection{Problem Formulation}

Multi-dimensional evaluation of NLG requires to evaluate $n$ particular dimensions $d=(d_1, \ldots, d_n)$ of the model output, and the input can include the candidate output $x$, reference text $y$, and context $c$.
$y$ is removed when evaluating reference-independent dimensions, such as \texttt{consistency} in summarization.
Depending on the specific generation task, $c$ can contain different content or even be omitted.
Evaluators need to evaluate the quality of the model output on each dimension and output scores $s=(s_1, \ldots, s_n)$ for all the dimensions.

To unify all evaluation dimensions into one evaluator, we transform each dimension into a Boolean question $q_i$.
For example, for $d_i$ = \texttt{coherence} in summarization, the transformed question $q_i$ is \textit{``Is this a coherent summary to the document?''}.
Then for each input $(x, y, c, q_i)$, evaluator should output ``Yes'' or ``No'' and calculate $s_i$ as:

\begin{align}
    s_i = \frac{P(\mathrm{``Yes"}|x, y, c, q_i)}{P(\mathrm{``Yes"}|x, y, c, q_i)+P(\mathrm{``No"}|x, y, c, q_i)},
\label{equ:score}
\end{align}

\noindent where $P(\mathrm{\cdot})$ denotes the probability of the model generating a specific word.
In this way, a single evaluator can evaluate $x$ of all dimensions by modifying the question description.

\subsection{Unsupervised Learning on Multiple Evaluation Dimensions}
\label{sec:evaluation_task}

Since annotating large-scale human scores to judge the quality of the generated text is unaffordable, we adopt an unsupervised setting to develop our evaluator.
Using T5~\cite{raffel2020exploring} as the backbone model, we first design specific rules for several commonly evaluated dimensions to construct pseudo data, and then combine them to train the evaluator.

\paragraph{Pseudo Data Construction} To train an evaluator, we need to construct positive and negative samples for different dimensions.
The former implies high-quality generated text, so we use groundtruth such as the reference summary in summarization.
Then we propose particular rules for each dimension to convert positive samples into negative ones.

Taking text summarization as an example, the specific rule-based transformations are as follows:

\noindent 1) \textbf{Coherence} refers to whether all the sentences form a coherent body.
To build incoherent summaries, we use BM25~\cite{robertson2009probabilistic} to retrieve similar summaries, and randomly select a sentence from the retrieved summary to replace one of the sentences in groundtruth summary.

\noindent 2) \textbf{Consistency} is the factual alignment between the summary and the source document.
We use the method in \citet{chen2021factuality} to construct inconsistent summaries by antonym substitution, numerical editing, entity replacement, and syntactic pruning.

\noindent 3) \textbf{Fluency} represents the quality of individual sentences. We randomly draw a span\footnote{The length of the span is sampled from the Poisson distribution ($\lambda=5$).} from the positive sample and perform one of repeating, deleting, and shuffling to obtain the disfluent summaries.

\noindent 4) \textbf{Relevance} means whether the summary contains only the important information of the source document.
The transformation rule is similar to coherence, except that we replace multiple sentences at random instead of one.

We include the designed rules for other NLG tasks in Appendix~\ref{sec:pseudo_dialog}.
The detailed descriptions and concrete examples for all  dimensions can also be found in Appendices~\ref{sec:explanation} and \ref{sec:example_evaluation}.

\paragraph{Training Strategy} 
For each generation task, we attempt to build a single evaluator to evaluate the NLG model from different dimensions.
A straightforward approach is to perform multi-task learning on synthetic data of all dimensions to obtain a unified evaluator.
However, we observe the negative transfer problem in several dimensions (e.g., \texttt{coherence} in summarization and \texttt{engagingness} in dialogue generation, see Tables~\ref{tab:summarization} and \ref{tab:dialogue}).
To tackle this issue, we employ a simple and effective method from continual learning~\cite{parisi2019continual}: whenever a new dimension is introduced, we add small portion of data from all previous dimensions to replay.
The benefit is that we can easily extend our evaluator to new dimensions without training from scratch.
Moreover, this method enables to explicitly learn dimensions related to linguistic features (e.g., \texttt{fluency}) first, and then move on to the dimensions that require a better understanding of the text (e.g., \texttt{consistency}).
We show that this sequential training approach can alleviate the negative transfer problem in Section~\ref{sec:exp}.

\subsection{Intermediate Multi-task Learning}
Benefiting from the unified Boolean QA format, we can additionally introduce intermediate tasks for \textsc{UniEval}  to incorporate external knowledge from existing related datasets.
As shown in Figure~\ref{fig:model}, this stage is placed before the unsupervised learning on evaluation tasks.
Notably, the input here is $(c, q)$, which no longer includes the candidate output $x$ and the reference text $y$.
In total, we collect four types of intermediate tasks as follows.

\noindent \textbf{Natural Language Inference.} The task of NLI is to determine whether a ``hypothesis'' is true (entailment), false (contradiction), or undetermined (neutral) under a ``premise''.
We transform the NLI task into a question: ``\textit{Is this hypothesis entailed in the premise?}'', and only convert entailment into the label ``Yes'' and the rest to ``No''.
The context $c$ consists of a hypothesis and a premise.
We use the following three datasets: document-level NLI~\cite{yin2021docnli}, MRPC corpus~\cite{dolan2005automatically} and QQP~\cite{wang2017bilateral}.

\noindent \textbf{Self-Supervised Task.} Based on the classical next sentence prediction task~\cite{devlin2019bert}, we propose a new \emph{opening sentence prediction task}.
The goal of this task is to determine whether a sentence is the starting sentence of a given news article.
The motivation is that the first few sentences in the news tend to be salient and informative~\cite{see2017get, zhong2019searching}, so it allows the model to learn inter-sentence coherence while also capturing the central idea of the document.
We sample news from the CNN/DailyMail news corpus~~\cite{hermann2015teaching} and randomly select the opening sentence of other news as negative samples.

\noindent \textbf{Linguistics-Related Task.} To facilitate the incorporation of linguistic knowledge into the unified model, we also include CoLA dataset~\cite{warstadt2019neural} as the linguistic task.
This requires the model to judge whether a sentence is linguistically acceptable, so the input question is: ``\textit{Is this a fluent and linguistically acceptable sentence?}''.

\begin{table}[t]
\center \footnotesize
\tabcolsep0.09 in
\begin{tabular}{lccc}
\toprule
\textbf{Tasks} & \multicolumn{1}{c}{\textbf{\# Positive}} &  \multicolumn{1}{c}{\textbf{\# Negative}} & \multicolumn{1}{c}{\textbf{Total}} \\

\midrule

NLI & 41,149 & 44,652 & 85,801 \\
Self-supervised & 30,000 & 30,000 & 60,000 \\
Linguistics & 6,744 & 2,850 & 9,594 \\
Generic QA & 17,032 & 13,096 & 30,128 \\

\midrule
All & 94,925 & 90,598 & 185,523 \\

\bottomrule
\end{tabular}
\caption{Statistics for intermediate tasks. Positive sample indicates the model should answer with ``Yes''.}
\label{tab:statistics}
\end{table}

\noindent \textbf{Generic QA.} We collect the existing Boolean QA datasets: BoolQ~\cite{clark2019boolq}, BoolQ-NP~\cite{khashabi2020more}, BoolQ-CS~\cite{gardner2020evaluating},  StrategyQA~\cite{geva2021did}, and extract the questions in MultiRC dataset~\cite{khashabi2018looking} that can be answered with Yes/No as the data for generic QA task. 
Introducing these diverse question descriptions enables the model to better understand the importance of \textit{question} in the input format as well as incorporate more open-ended external knowledge.

The statistics of data can be found in Table~\ref{tab:statistics} and concrete examples for each task are also provided in Appendix~\ref{sec:example_intermediate}.
Since this phase is not related to the evaluation metric, we train the model with cross-entropy loss without computing $s_i$.

\section{Experiments}
\label{sec:exp}
Following~\citet{deng2021compression}, we classify NLG tasks into three types: compression, creation, and transduction, and select typical tasks from each category to conduct experiments.
For compression and creation, we choose summarization and dialogue response generation to measure the performance of \textsc{UniEval}, as well as the ability to zero-shot to unseen dimensions.
For transduction, we select data-to-text to test whether \textsc{UniEval} has the ability to transfer to a new NLG task.

\subsection{Implementation Details}
We use ``google/t5-v1\_1-large'' version of T5 as the backbone model in all the experiments.
The number of pseudo samples for each dimension is 30k, with an equal number of positive and negative examples.
The order for continual learning is \texttt{coherence} $\rightarrow$ \texttt{fluency} $\rightarrow$ \texttt{consistency} $\rightarrow$ \texttt{relevance} for summarization, and \texttt{coherence} $\rightarrow$ \texttt{naturalness} $\rightarrow$ \texttt{groundedness} $\rightarrow$ \texttt{engagingness} for dialogue generation.
For the score calculation, we follow previous work to compute sentence-level average scores for \texttt{fluency} and \texttt{consistency}~\cite{laban2021summac} in summarization, and sentence-level cumulative scores for \texttt{engagingness}~\cite{deng2021compression}, while the rest is calculated as Equation~\ref{equ:score}. More details can be found in Appendix~\ref{sec:implementation}.

\renewcommand\arraystretch{1.1}
\begin{table*}[t]
\center \footnotesize
\tabcolsep0.1 in
\begin{tabular}{lcccccccccc}
\toprule
\multicolumn{1}{l}{\multirow{2}[1]{*}{\textbf{Metrics}}} & \multicolumn{2}{c}{\textbf{Coherence}}
 & \multicolumn{2}{c}{\textbf{Consistency}} & \multicolumn{2}{c}{\textbf{Fluency}} & \multicolumn{2}{c}{\textbf{Relevance}} & \multicolumn{2}{c}{\textbf{Average}} \\
 & \multicolumn{1}{c}{$\rho$} & \multicolumn{1}{c}{$\tau$} & \multicolumn{1}{c}{$\rho$} & \multicolumn{1}{c}{$\tau$} & \multicolumn{1}{c}{$\rho$} & \multicolumn{1}{c}{$\tau$} & \multicolumn{1}{c}{$\rho$} & \multicolumn{1}{c}{$\tau$} & \multicolumn{1}{c}{$\rho$} & \multicolumn{1}{c}{$\tau$}  \\
 
\cmidrule(lr){1-1} \cmidrule(lr){2-3} \cmidrule(lr){4-5} \cmidrule(lr){6-7} \cmidrule(lr){8-9} \cmidrule(lr){10-11}

\multicolumn{11}{l}{\textbf{Similarity-based Metrics}} \\

\textsc{ROUGE-1} & 0.167 & 0.126 & 0.160 & 0.130 & 0.115 & 0.094 & 0.326 & 0.252 & \cellcolor[rgb]{ .851,  .851,  .851} 0.192 & \cellcolor[rgb]{ .851,  .851,  .851} 0.150 \\
\textsc{ROUGE-2} & 0.184 & 0.139 & 0.187 & 0.155 & 0.159 & 0.128 & 0.290 & 0.219 & \cellcolor[rgb]{ .851,  .851,  .851} 0.205 & \cellcolor[rgb]{ .851,  .851,  .851} 0.161 \\
\textsc{ROUGE-L} & 0.128 & 0.099 & 0.115 & 0.092 & 0.105 & 0.084 & 0.311 & 0.237 & \cellcolor[rgb]{ .851,  .851,  .851} 0.165 & \cellcolor[rgb]{ .851,  .851,  .851} 0.128 \\
\textsc{BertScore} & 0.284 & 0.211 & 0.110 & 0.090 & 0.193 & 0.158 & 0.312 & 0.243 & \cellcolor[rgb]{ .851,  .851,  .851} 0.225 & \cellcolor[rgb]{ .851,  .851,  .851} 0.175 \\
\textsc{MoverScore} & 0.159 & 0.118 & 0.157 & 0.127 & 0.129 & 0.105 & 0.318 & 0.244 & \cellcolor[rgb]{ .851,  .851,  .851} 0.191 & \cellcolor[rgb]{ .851,  .851,  .851} 0.148 \\

\midrule
\multicolumn{11}{l}{\textbf{Single-dimensional Evaluators}} \\

\textsc{CTC} (Consistency) & \underline{0.223} & \underline{0.172} & 0.415 & 0.345 & \underline{0.335} & \underline{0.276} & \underline{0.166} & \underline{0.124} & \cellcolor[rgb]{ .851,  .851,  .851} 0.285 & \cellcolor[rgb]{ .851,  .851,  .851} 0.229 \\

\textsc{CTC} (Relevance) & \underline{0.402} & \underline{0.310} & \underline{0.366} & \underline{0.301} & \underline{0.299} & \underline{0.245} & 0.428 & 0.336 & \cellcolor[rgb]{ .851,  .851,  .851} 0.374 & \cellcolor[rgb]{ .851,  .851,  .851} 0.298 \\

\textsc{UniEval} (Coherence) & \textbf{0.546} & \textbf{0.422} & \underline{0.337} & \underline{0.280} & \underline{0.324} & \underline{0.266} & \underline{0.418} & \underline{0.316} & \cellcolor[rgb]{ .851,  .851,  .851} 0.406 & \cellcolor[rgb]{ .851,  .851,  .851} 0.321 \\

\textsc{UniEval} (Consistency) & \underline{0.176} & \underline{0.127} & \textbf{0.472} & \textbf{0.393} & \underline{0.366} & \underline{0.300} & \underline{0.176} & \underline{0.128} & \cellcolor[rgb]{ .851,  .851,  .851} 0.298 & \cellcolor[rgb]{ .851,  .851,  .851} 0.237 \\

\textsc{UniEval} (Fluency) & \underline{0.324} & \underline{0.247} & \underline{0.276} & \underline{0.229} & \textbf{0.433} & \textbf{0.360} & \underline{0.236} & \underline{0.176} & \cellcolor[rgb]{ .851,  .851,  .851} 0.317 & \cellcolor[rgb]{ .851,  .851,  .851} 0.253 \\

\textsc{UniEval} (Relevance) & \underline{0.543} & \underline{0.420} & \underline{0.324} & \underline{0.267} & \underline{0.340} & \underline{0.283} & \textbf{0.463} & \textbf{0.355} & \cellcolor[rgb]{ .851,  .851,  .851} 0.417 & \cellcolor[rgb]{ .851,  .851,  .851} 0.332 \\

\midrule
\multicolumn{11}{l}{\textbf{Unified Evaluators}} \\

\textsc{BartScore} & 0.448 & 0.342 & 0.382 & 0.315 & 0.356 & 0.292 & 0.356 & 0.273 & \cellcolor[rgb]{ .851,  .851,  .851} 0.385 & \cellcolor[rgb]{ .851,  .851,  .851} 0.305 \\

\textsc{UniEval} (Multi-task) & 0.495 & 0.374 & 0.435 & 0.365 & 0.419 & 0.346 & 0.424 & \textbf{0.327} &  \cellcolor[rgb]{ .851,  .851,  .851} 0.443 & \cellcolor[rgb]{ .851,  .851,  .851} 0.353 \\

\textsc{UniEval} (Continual) & \textbf{0.575} & \textbf{0.442} & \textbf{0.446} & \textbf{0.371} & \textbf{0.449} & \textbf{0.371} & \textbf{0.426} & 0.325 & \cellcolor[rgb]{ .851,  .851,  .851} \textbf{0.474} & \cellcolor[rgb]{ .851,  .851,  .851} \textbf{0.377} \\

\quad - Intermediate Tasks & 0.477 & 0.363 & 0.403 & 0.333 & 0.414 & 0.342 & 0.395 & 0.301 & \cellcolor[rgb]{ .851,  .851,  .851} 0.422 & \cellcolor[rgb]{ .851,  .851,  .851} 0.335 \\

\bottomrule
\end{tabular}
\caption{Summary-level Spearman ($\rho$) and Kendall-Tau ($\tau$) correlations of different metrics on SummEval benchmark. The underlined numbers indicate the results of transferring a single-dimensional evaluator to other dimensions.}
\label{tab:summarization}
\end{table*}

\subsection{Baselines}

We compare \textsc{UniEval} with several state-of-the-art evaluators.
Notably, all the single-dimensional and unified evaluators are built on the same corpus.

\textbf{BERTScore}~\cite{zhang2019bertscore} is a similarity-based evaluator. It computes the similarity between two text sequences based on the contextualized embedding obtained by BERT~\cite{devlin2019bert}.

\textbf{MoverScore}~\cite{zhao2019moverscore} adds many-to-one alignment to BERTScore and introduces new aggregation methods to achieve a more powerful similarity-based evaluator.

\textbf{CTC}~\cite{deng2021compression} utilizes information alignment to define metrics for several specific dimensions in NLG tasks, and proposes three model variants for each dimension. We compare the best variants of CTC in each dimension as the single-dimensional evaluators in our experiments.

\textbf{BARTScore}~\cite{yuan2021bartscore} is a unified evaluator which uses average likelihood of the model output as the metric. It can predict different scores depending on the different input and output. We follow the original paper using $c \rightarrow x$ as the score for \texttt{coherence}, \texttt{consistency} and \texttt{fluency}, and $x \rightarrow y$ as the score for \texttt{relevance}.

\textbf{USR}~\cite{mehri2020usr} is a unified evaluator designed for dialogue response generation task. It uses different variants (e.g., MLM, dialogue retrieval and overall metric) to predict multiple scores for each generated response. We choose the score with the best correlation for each dimension for comparison in the experiments.

\subsection{Benchmarks}

We adopt four meta-evaluation benchmarks for various NLG tasks to measure the correlation between \textsc{UniEval} and human judgments.

\textbf{SummEval}~\cite{fabbri2021summeval} is a meta-evaluation benchmark for summarization. For each summary to be evaluated, it provides human scores from four dimensions: \texttt{fluency}, \texttt{coherence}, \texttt{consistency} and \texttt{relevance}. We use it to measure the performance of \textsc{UniEval}.

\textbf{Topical-Chat}~\cite{mehri2020usr} is a benchmark for knowledge-based dialogue response generation task. It includes human scores from five dimensions: \texttt{naturalness}, \texttt{coherence}, \texttt{engagingness}, \texttt{groundedness} and \texttt{understandability}\footnote{We rephrase (natural, maintains context, interesting, uses knowledge, understandable) from the original paper into the five dimensions mentioned above.}. The first four dimensions are used to measure the performance of \textsc{UniEval}, and the last one is used for the transfer experiment.

\textbf{SFRES} and \textbf{SFHOT}~\cite{wen2015semantically} are meta-evaluation benchmarks for data-to-text task. They provide information about restaurants and hotels in San Francisco and aim to let the model generate corresponding utterances. We leverage the annotations of \texttt{informativeness} and \texttt{naturalness} dimension to conduct transfer experiment.

\textbf{QAGS}~\cite{wang2020asking} is also a benchmark for summarization. It is designed to detect \texttt{consistency} dimension on two summarization corpora~\cite{narayan2018don}. We use it to test the performance of the single-dimensional version of \textsc{UniEval}, and the results are listed in Appendix~\ref{sec:qags}.

\renewcommand\arraystretch{1.1}
\begin{table*}[t]
\center \footnotesize
\tabcolsep0.095 in
\begin{tabular}{lcccccccccc}
\toprule
\multicolumn{1}{l}{\multirow{2}[1]{*}{\textbf{Metrics}}} & \multicolumn{2}{c}{\textbf{Naturalness}}
 & \multicolumn{2}{c}{\textbf{Coherence}} & \multicolumn{2}{c}{\textbf{Engagingness}} & \multicolumn{2}{c}{\textbf{Groundedness}} & \multicolumn{2}{c}{\textbf{Average}} \\
  & \multicolumn{1}{c}{$r$} & \multicolumn{1}{c}{$\rho$} & \multicolumn{1}{c}{$r$} & \multicolumn{1}{c}{$\rho$}  & \multicolumn{1}{c}{$r$} & \multicolumn{1}{c}{$\rho$}  & \multicolumn{1}{c}{$r$} & \multicolumn{1}{c}{$\rho$}  & \multicolumn{1}{c}{$r$} & \multicolumn{1}{c}{$\rho$}   \\
 
\cmidrule(lr){1-1} \cmidrule(lr){2-3} \cmidrule(lr){4-5} \cmidrule(lr){6-7} \cmidrule(lr){8-9} \cmidrule(lr){10-11}

\multicolumn{11}{l}{\textbf{Similarity-based Metrics}} \\

\textsc{BLEU-1} & 0.161 & 0.133 & 0.210 & 0.223 & 0.314 & 0.334 & 0.289 & 0.303 & \cellcolor[rgb]{ .851,  .851,  .851} 0.243 & \cellcolor[rgb]{ .851,  .851,  .851} 0.248 \\

\textsc{BLEU-4} & 0.180 & 0.175 & 0.131 & 0.235 & 0.232 & 0.316 & 0.213 & 0.310 & \cellcolor[rgb]{ .851,  .851,  .851} 0.189 & \cellcolor[rgb]{ .851,  .851,  .851} 0.259 \\

\textsc{ROUGE-L} & 0.176 & 0.146 & 0.193 & 0.203 & 0.295 & 0.300 & 0.310 & 0.327 & \cellcolor[rgb]{ .851,  .851,  .851} 0.243 & \cellcolor[rgb]{ .851,  .851,  .851} 0.244 \\

\textsc{METEOR} & 0.212 & 0.191 & 0.250 & 0.302 & 0.367 & 0.439 & 0.333 & 0.391 & \cellcolor[rgb]{ .851,  .851,  .851} 0.290 & \cellcolor[rgb]{ .851,  .851,  .851} 0.331 \\

\textsc{BertScore} & 0.226 & 0.209 & 0.214 & 0.233 & 0.317 & 0.335 & 0.291 & 0.317 & \cellcolor[rgb]{ .851,  .851,  .851} 0.262 & \cellcolor[rgb]{ .851,  .851,  .851} 0.273 \\

\midrule
\multicolumn{11}{l}{\textbf{Single-dimensional Evaluators}} \\

\textsc{CTC} (Engagingness) & \underline{0.280} & \underline{0.257} & \underline{0.352} & \underline{0.325} & 0.516 & 0.525 & \underline{0.405} & \underline{0.404} & \cellcolor[rgb]{ .851,  .851,  .851} 0.388 & \cellcolor[rgb]{ .851,  .851,  .851} 0.378 \\

\textsc{CTC} (Groundedness) & \underline{0.200} & \underline{0.161} & \underline{0.256} & \underline{0.228} & \underline{0.485} & \underline{0.475} & 0.524 & 0.477 & \cellcolor[rgb]{ .851,  .851,  .851} 0.366 & \cellcolor[rgb]{ .851,  .851,  .851} 0.335 \\

\textsc{UniEval} (Naturalness) & \textbf{0.500} & \textbf{0.547} & \underline{0.331} & \underline{0.458} & \underline{0.393} & \underline{0.528} & \underline{0.178} & \underline{0.266} & \cellcolor[rgb]{ .851,  .851,  .851} 0.351 & \cellcolor[rgb]{ .851,  .851,  .851} 0.450 \\

\textsc{UniEval} (Coherence) & \underline{0.401} & \underline{0.468} & \textbf{0.543} & \textbf{0.607} & \underline{0.401} & \underline{0.474} & \underline{0.225} & \underline{0.235} & \cellcolor[rgb]{ .851,  .851,  .851} 0.392 & \cellcolor[rgb]{ .851,  .851,  .851} 0.446 \\

\textsc{UniEval} (Engagingess) & \underline{0.394} & \underline{0.427} & \underline{0.471} & \underline{0.477} & \textbf{0.562} & \textbf{0.596} & \underline{0.376} & \underline{0.431} & \cellcolor[rgb]{ .851,  .851,  .851} 0.451 & \cellcolor[rgb]{ .851,  .851,  .851} 0.483 \\

\textsc{UniEval} (Groundedness) & \underline{0.220} & \underline{0.153} & \underline{0.187} & \underline{0.117} & \underline{0.392} & \underline{0.318} & \textbf{0.543} & \textbf{0.511} & \cellcolor[rgb]{ .851,  .851,  .851} 0.336 & \cellcolor[rgb]{ .851,  .851,  .851} 0.275 \\

\midrule
\multicolumn{11}{l}{\textbf{Unified Evaluators}} \\

\textsc{USR} & 0.337 & 0.325 & 0.416 & 0.377 & 0.456 & 0.465 & 0.222 & 0.447 & \cellcolor[rgb]{ .851,  .851,  .851} 0.358 & \cellcolor[rgb]{ .851,  .851,  .851} 0.403 \\

\textsc{UniEval} (Multi-task) & \textbf{0.480} & 0.512 & 0.518 & 0.609 & 0.544 & 0.563 & 0.462 & 0.456 & \cellcolor[rgb]{ .851,  .851,  .851} 0.501 & \cellcolor[rgb]{ .851,  .851,  .851} 0.535 \\

\textsc{UniEval} (Continual) & 0.444 & \textbf{0.514} & \textbf{0.595} & \textbf{0.613} & \textbf{0.557} & \textbf{0.605} & \textbf{0.536} & \textbf{0.575} & \cellcolor[rgb]{ .851,  .851,  .851} \textbf{0.533} & \cellcolor[rgb]{ .851,  .851,  .851} \textbf{0.577} \\

\quad - Intermediate Tasks & 0.442 & 0.478 & 0.532 & 0.579 & 0.537 & 0.555 & 0.410 & 0.440 & \cellcolor[rgb]{ .851,  .851,  .851} 0.480 & \cellcolor[rgb]{ .851,  .851,  .851} 0.513 \\

\bottomrule
\end{tabular}
\caption{Turn-level Pearson ($r$) an Spearman ($\rho$) correlations of different metrics on the Topical-Chat benchmark. The underlined numbers indicate the results of transferring a single-dimensional evaluator to other dimensions.}
\label{tab:dialogue}
\end{table*}

\subsection{Results For Summarization}
Following \citet{liu2021explainaboard}, we use summary-level Spearman and Kendall-Tau correlation to assess the performance of different evaluators for summarization.
Results of similarity-based metrics are listed in the first part of Table~\ref{tab:summarization}.
They are designed to measure the semantic overlap between the model output and the reference text, so they can obtain relatively high correlations in \texttt{relevance} dimension.
However, they are not qualified metrics for the other dimensions due to the poor correlation.

The second part contains the results of single-dimensional evaluators.
CTC is currently the best evaluators of \texttt{consistency} and \texttt{relevance}, but it fails to excel on \texttt{coherence} and \texttt{fluency}.
Here we also adapt \textsc{UniEval} to several single-dimensional variants by training the model on pseudo data of only one dimension.
Our proposed evaluators exceed  CTC models and achieve the best correlation in all dimensions.
It reveals that our proposed Boolean QA formulation can clearly enhance the backbone pre-trained model.
Furthermore, we attempt to transfer the single-dimensional evaluators to other dimensions, and the underlined numbers in Table~\ref{tab:summarization} are transferred results.
Overall \textsc{UniEval} is better than CTC, but we can see that no single-dimensional evaluator can transfer well to all dimensions.
For example, both \texttt{consistency} $\Rightarrow$ \texttt{coherence} and \texttt{fluency} $\Rightarrow$ \texttt{relevance} exhibit poor correlations, indicating that evaluators that focus solely on a single evaluation dimension lack acceptable transfer capability.

As shown in the last part, \textsc{UniEval} substantially surpasses the state-of-the-art unified evaluator BARTScore in the summarization task.
Specifically, \textsc{UniEval} trained by multi-task learning brings an average improvement of more than 15\% across all dimensions compared to BARTScore.
And this gain is boosted to more than 23\%  by adapting continual learning in the unsupervised learning phase.
The main gap between the two training strategies of \textsc{UniEval} is the negative transfer on \texttt{coherence}, which clarifies that explicitly learning basic language features before learning more complex dimensions can alleviate this problem.
It is also notable that compared with its single-dimensional version, the unified version of \textsc{UniEval} is improved in both \texttt{coherence} and \texttt{fluency}, while having a slight decrement in the other two dimensions.
This suggests that following continual learning, we can sequentially extend our evaluator to a new dimension while preserving the performance on previous dimensions.
Moreover, the clear performance drop after removing the intermediate tasks in the last row illustrates the importance and usefulness of this phase.

\subsection{Results For Dialogue Generation}

To test the performance of different evaluators on the dialogue response generation task, we compute turn-level Pearson and Spearman correlation on the Topical-chat benchmark as in \citet{mehri2020usr}.
Table~\ref{tab:dialogue} presents that similarity-based metrics correlate relatively well on \texttt{engagingness} and \texttt{groundedness} while performing poorly on the remaining dimensions.
With respect to the single-dimensional evaluator, we can reach the same conclusion as for the summarization task: the scores predicted by \textsc{UniEval} have the highest correlation with human judgments in all dimensions. 

Compared to USR, the state-of-the-art unified evaluator in the dialogue response generation task, our evaluator demonstrates more remarkable boosts.
According to Pearson and Spearman correlation, \textsc{UniEval} (Continual) improves the results by an average of 48.9\% and 43.2\%, respectively.
In comparison with the corresponding single-dimensional version, although there is a performance loss in \texttt{naturalness},  \textsc{UniEval} (Continual) brings improvements in the remaining dimensions based on Spearman correlation.
Especially for \texttt{grounedness}, the unified version increases the correlation by 12.5\% (0.511 $\Rightarrow$ 0.575) compared to the single-dimensional version.
Meanwhile, intermediate tasks also display an indispensable role in evaluating dialogue generation, indicating that its benefits can span a variety of NLG tasks.

\subsection{Transfer Experiments}

We perform two zero-shot experiments to exhibit the transfer ability of \textsc{UniEval}.

\begin{figure}
    \centering
    \includegraphics[width=1.0\linewidth]{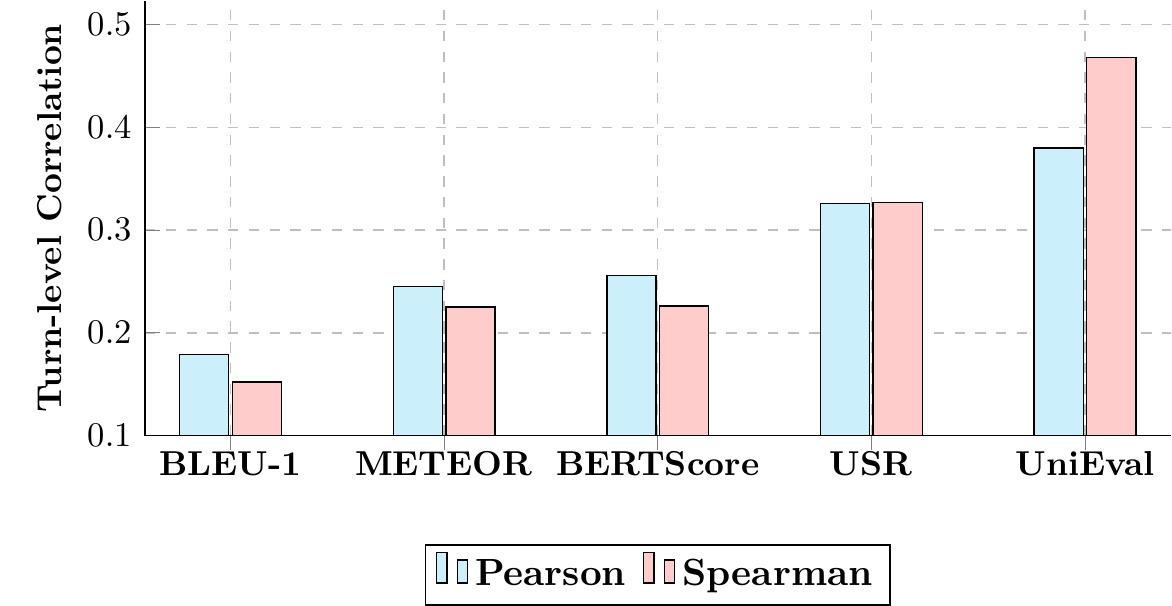}
    \caption{Zero-shot performance on the ``understandability'' dimension in dialogue response generation.}
    \label{fig:understandability}
\end{figure}

\paragraph{Zero-shot to Unseen Dimension} 
To meet the requirements of different users, new evaluation dimensions often emerge for particular NLG tasks.
For instance, certain users may prefer a new \texttt{``understandability''} dimension over other dimensions for the dialogue generation task.
Therefore, we conduct experiments on the Topical-chat meta-evaluation benchmark to observe if \textsc{UniEval}\footnote{Here is \textsc{UniEval} (Continual) in Table~\ref{tab:dialogue}.} has the transfer capability in this scenario.
Concretely, we adjust the input question to \textit{``Is this an understandable response in the dialogue?''}, and calculate the metric based on Equation~\ref{equ:score}.
As shown in Figure~\ref{fig:understandability}, although \textsc{UniEval} has not seen or been trained on this dimension before, its predicted score still correlates well with human judgments.
It even outperforms the best USR metric for both Pearson (0.326 $\Rightarrow$ 0.380) and Spearman (0.327 $\Rightarrow$ 0.468) correlations, which denotes that \textsc{UniEval} is capable of transferring to unseen dimensions by modifying the prompt.

\paragraph{Zero-shot to Unseen Task}
In a more radical setting, we also transfer \textsc{UniEval} to a new NLG task of data-to-text generation in the zero-shot setting.
As annotated in the SFRES and SFHOT benchmarks, users emphasize the \texttt{naturalness} and \texttt{informativeness} of the generated utterance for this task.
Therefore, we adapt the question to ``\textit{Is this a fluent utterance?}'' and ``\textit{Is this sentence informative according to the reference?}'' to predict the evaluation scores for these two dimensions.
``T5 + intermediate'' in Table~\ref{tab:d2t} represents the model obtained after the intermediate multi-learning stage.
While it has not been trained on any evaluation tasks, it performs on par with BARTScore based on average correlations and is particularly good at evaluating the \texttt{naturalness} of utterances.
After training on multiple evaluation dimensions on summarization, \textsc{UniEval} (Summ)\footnote{Here is equivalent to \textsc{UniEval} (Continual) in Table~\ref{tab:summarization}.} demonstrates better transfer ability and superior performance over BARTScore in most dimensions of both datasets.
This illustrates the capability of \textsc{UniEval} to transfer to new NLG tasks without further adaptation. 

\begin{table}[t]
\center \footnotesize
\tabcolsep0.05 in
\begin{tabular}{l|cc|cc|c}
\toprule
\multicolumn{1}{l}{\multirow{2}[1]{*}{\textbf{Metrics}}} & \multicolumn{2}{c}{\textbf{SFRES}} & 
\multicolumn{2}{c}{\textbf{SFHOT}} &
\multicolumn{1}{c}{\multirow{2}[1]{*}{\textbf{Avg.}}} \\
 & \textbf{Nat.} & \textbf{Info.} & \textbf{Nat.} & \textbf{Info.} \\

\midrule

ROUGE-1 & 0.170 & 0.115 & 0.196 & 0.118 & \cellcolor[rgb]{ .851,  .851,  .851} 0.150 \\
ROUGE-L & 0.169 & 0.103 & 0.186 & 0.110 & \cellcolor[rgb]{ .851,  .851,  .851} 0.142 \\
\textsc{BertScore} & 0.219 & 0.156 & 0.178 & 0.135 & \cellcolor[rgb]{ .851,  .851,  .851} 0.172 \\
\textsc{MoverScore} & 0.190 & 0.153 & 0.242 & 0.172 & \cellcolor[rgb]{ .851,  .851,  .851} 0.189 \\
\textsc{BARTScore} & 0.289 & \textbf{0.238} & 0.288 & 0.235 & \cellcolor[rgb]{ .851,  .851,  .851} 0.263 \\

\midrule
T5 + Intermediate & \textbf{0.348} & 0.180 & 0.310 & 0.181 & \cellcolor[rgb]{ .851,  .851,  .851} 0.255 \\
\textsc{UniEval} (Summ) & 0.333 & 0.225 & \textbf{0.320} & \textbf{0.249} & \cellcolor[rgb]{ .851,  .851,  .851} \textbf{0.282} \\

\bottomrule
\end{tabular}
\caption{Zero-shot performance (Spearman) on the data-to-text task. Nat. and Info. denote \texttt{Naturalness} and \texttt{Informativeness}, respectively.}
\label{tab:d2t}
\end{table}

\begin{table}[t]
\center \footnotesize
\tabcolsep0.06 in
\begin{tabular}{lccccc}
\toprule
\textbf{Evaluators} & \textbf{Coh.} & \textbf{Con.} & \textbf{Flu.} & \textbf{Rel.} & \textbf{Avg.} \\

\midrule

\textsc{UniEval} & 0.546 & \textbf{0.472} & 0.433 & \textbf{0.463} & \cellcolor[rgb]{ .851,  .851,  .851} \textbf{0.479} \\

\quad - NLI & 0.532 & 0.417 & \textbf{0.436} & 0.452 & \cellcolor[rgb]{ .851,  .851,  .851} 0.459 \\

\quad - SST & 0.498 & 0.462 & 0.428 & 0.450 & \cellcolor[rgb]{ .851,  .851,  .851} 0.460 \\

\quad - Linguistics & \textbf{0.548} & 0.466 & 0.415 & 0.458 & \cellcolor[rgb]{ .851,  .851,  .851} 0.472 \\

\quad - Generic QA & 0.528 & 0.438 & 0.421 & 0.436 & \cellcolor[rgb]{ .851,  .851,  .851} 0.456 \\

\bottomrule
\end{tabular}
\caption{Ablation study of \textsc{UniEval}. ‘-’ means we remove this task from interemediate multi-task learning.}
\label{tab:ablation}
\end{table}

\subsection{Ablation Study of Intermediate Tasks}
We conduct ablation studies on the single-dimensional version of \textsc{UniEval} to better investigate the contribution of each type of intermediate task to NLG evaluation.
The results of Spearman correlation are presented in Table~\ref{tab:ablation}.
Because of the similar task requirements, NLI contributes most to \texttt{consistency}, while our proposed opening sentence prediction task facilitates the evaluator to capture \texttt{coherence} between sentences.
Due to the small data size of the linguistics-related task (see Table~\ref{tab:statistics}), removing it does not have a significant impact on the performance, but it can still help the model better understand \texttt{fluency} of individual sentences.
Generic QA enhances each dimension by engaging the evaluator to focus on the meaning of the input question.
Overall, training on the combination of all four types of intermediate tasks leads to the best NLG evaluation performance.

\section{Conclusion}
In this paper, we emphasize the necessity of multi-dimensional evaluation in advancing the field of NLG.
To promote this comprehensive and fine-grained evaluation approach, we propose a unified multi-dimensional evaluator \textsc{UniEval} for various NLG tasks.
\textsc{UniEval} correlates well with human judgment on three typical generation tasks and exhibits excellent transfer performance.

\section*{Limitations}

We state the limitations of this paper from the following four aspects:

1) Most of the current evaluators, including \textsc{UniEval}, are black-box models.
With the support of pre-trained language model, even though the neural evaluators can already correlate well with human judgments, it is still unclear how the model predicts these evaluation scores.
Therefore, a better understanding of the evaluation process of different evaluators or the development of an interpretable and multi-dimensional evaluator may be the next stage for improving NLG evaluation.

2) Most of the neural evaluators are trained on synthetic data, while the pseudo data constructed in this paper still contain noise.
For instance, for \texttt{fluency} in summarization, removing an unimportant span may not affect the fluency of the sentence, but we always treat the sentence after deleting as a negative sample.
Thus, how to improve the quality of synthetic data could be an interesting topic.

3) We only use T5-large as the backbone model in the experiments due to the limited computational resources.
How to extend the use of neural evaluators by using smaller models but retaining similar performance, or how to introduce more data to build larger evaluators with better performance, could be two future research directions.

4) We follow the categorization of NLG tasks in \citet{deng2021compression} and select three typical tasks for our experiments, but \textsc{UniEval} is still limited to English tasks.
The generation tasks for cross-language scenarios are left for our future work.

\section*{Acknowledgements}
We thank Weizhe Yuan, Mingkai Deng, Yu Meng, Hou Pong Chan, Dan Iter and Reid Pryzant for helpful discussions and feedback.
We would also like to thank anonymous reviewers for valuable comments and suggestions.
Research was supported in part by US DARPA KAIROS Program No. FA8750-19-2-1004 and INCAS Program No. HR001121C0165, National Science Foundation IIS-19-56151, IIS-17-41317, and IIS 17-04532, and the Molecule Maker Lab Institute: An AI Research Institutes program supported by NSF under Award No. 2019897, and the Institute for Geospatial Understanding through an Integrative Discovery Environment (I-GUIDE) by NSF under Award No. 2118329. Any opinions, findings, and conclusions or recommendations expressed herein are those of the authors and do not necessarily represent the views, either expressed or implied, of DARPA or the U.S. Government. The views and conclusions contained in this paper are those of the authors and should not be interpreted as representing any funding agencies.

\bibliography{anthology,custom}
\bibliographystyle{acl_natbib}

\appendix

\clearpage
\section{Dimensions in Evaluation tasks}

\subsection{Explanation of Each Dimension}
\label{sec:explanation}

We introduce different dimensions for text summarization in Section~\ref{sec:evaluation_task}.
Here we include the detailed descriptions of different dimensions in dialogue response generation and data-to-text tasks.

For dialogue response generation~\cite{mehri2020usr}:

\begin{itemize}
    \item 1) \texttt{Naturalness}: judge whether a response is like something a person would naturally say
    \item 2) \texttt{Coherence}: determine whether this response serves as a valid continuation of the previous conversation.
    \item 3) \texttt{Engagingness}: determine if the response is interesting or dull.
    \item 4) \texttt{Groundedness}: given the fact that this response is conditioned on, determine whether this response uses that fact.
    \item 5) \texttt{Understandability}: judge whether the response is understandable.
\end{itemize}

For data-to-text~\cite{wen2015semantically}:

\begin{itemize}
    \item 1) \texttt{Naturalness}: determine whether the utterance could plausibly have been produced by a human.
    \item 2) \texttt{Informativeness}: determine whether the utterance contains all the information in the given content.
\end{itemize}

\subsection{Pseudo Data Construction for Dialogue Response Generation}
\label{sec:pseudo_dialog}

We produce pseudo data for the four dimensions of the dialogue response generation task as follows:

\begin{itemize}
    \item 1) \texttt{Naturalness}: similar to \texttt{fluency} in summarization, except that we modify $\lambda$ to 3.
    \item 2) \texttt{Coherence}: we randomly select gold response from other dialogues as the negative samples.
    \item 3) \texttt{Engagingness}: responses that are not engaging are dull and uninformative~\cite{mehri2020usr}. So we let DialogGPT-small~\cite{zhang2020dialogpt} generate response given just one sentence, thus creating unattractive samples.
    \item 4) \texttt{Groundedness}: this dimension is used to measure how well the response refers to the knowledge context in knowledge-based conversations~\cite{DBLP:conf/iclr/DinanRSFAW19}. 
Therefore, we randomly extract a sentence from the current knowledge context and use a paraphrase generator\footnote{\url{huggingface.co/tuner007/pegasus_paraphrase}} to rewrite it as a positive example, and sample a sentence from other knowledge contexts as a negative example.
\end{itemize}

\subsection{Examples for Evaluation Tasks}
\label{sec:example_evaluation}
\renewcommand\arraystretch{1.0}
\begin{table*}
\centering \footnotesize
\label{tab:pattern}
\begin{tabular}{lccc}
\toprule
\textbf{Dimensions} & \textbf{Tasks} & \textbf{Input} & \textbf{Target} \\
\midrule
Coherence & Summarization & \multicolumn{1}{m{9cm}}{\colorR{question}: Is this a coherent summary to the document? </s> \colorG{summary}: Theodore Wafer's statement contradicts his attorney's claim that he feared for his life and acted in self defense when he killed Renisha McBride. The shotgun is a Mossberg pump-action 12-gauge with a pistol grip. </s> \colorB{document}: On trial: Theodore Wafer, 55, initially told police that the shooting was an accident ...} & Yes \\
\midrule
Consistency & Summarization & 
\multicolumn{1}{m{9cm}}{\colorR{question}: Is this claim consistent with the document? </s> \colorG{claim}: The request wasn't rejected by Mr Justice Tugendhat last year and yesterday the Court of Appeal upheld that decision. </s> \colorB{document}: By James Slack The identity of suspects arrested by the police should be publicised before they are charged, the Court of Appeal has ruled ...} & No \\
\midrule
Fluency & Summarization &
\multicolumn{1}{m{9cm}}{\colorR{question}: Is this a fluent paragraph? </s> \colorG{paragraph}: Jack Bowlby's body discovered at Cheltenham College at Cheltenham College . He was described as a ``star pupil'' by staff at the £30,000-a-year school.} & No \\
\midrule
Relevance & Summarization & \multicolumn{1}{m{9cm}}{\colorR{question}: Is this summary relevant to the reference? </s> \colorG{summary}: The Met Office issued severe weather warnings across the country yesterday as experts predicted a weekend washout. Forecasters said wind and rain will continue to batter the country until Tuesday at the earliest, as fears grew that the Somerset Levels could be flooded again. The North will be wet and windy for the next three days, with showers also scattered across the South West. </s> \colorY{reference}: Heavy rain caused massive tailbacks yesterday as a pothole opened up across three lanes of the M25. Fear grow that the Somerset Levels could be flooded again. North will be wet and windy for next three days, with showers also scattered across the South West.} & Yes \\
\midrule
Naturalness & Dialogue &
\multicolumn{1}{m{9cm}}{\colorR{question}: Is this a natural response in the dialogue? </s> \colorG{response}: yes and that launched the career of many people, the most notable being han solo. the acting in the first was the most notable being han solo. the acting atrocious but got better as more movies were made. all told a great movie.} & No
\\
\midrule
Coherence & Dialogue &
\multicolumn{1}{m{9cm}}{\colorR{question}: Is this a coherent response given the dialogue history? </s> \colorG{response}: wow that is a lot of money for a logo. did you know corproate sponsors pay \$1.12 billion on the nba last year!? </s> \colorB{dialogue history}: hi! do you like basketball? \textbackslash n yes, i am a big raptors fan. it's crazy how much companies are paying to put their logos on nba jerseys. \textbackslash n i am sure it is extremely high to advertise on jerseys. do you know how much? \textbackslash n different team to team but everyone seems to get them these days. i did see geico is paying 6.5 million per year for their patch on the wizards jersey!  \textbackslash n\textbackslash n} & Yes
\\
\midrule
Engagingness & Dialogue &

\multicolumn{1}{m{9cm}}{\colorR{question}: Is this an engaging and informative response according to the dialogue history and fact? </s> \colorG{response}: i'm so glad i'm not the only one who thinks this. </s> \colorB{dialogue history}: do you follow american politics? \textbackslash n some,  i am not surprised that the first phone number in the white house was 1. lol \textbackslash n it definitely helped people reach the white house the fastest. i am surprised they still use floppy disks for storage. \textbackslash n\textbackslash n </s> \colorB{fact}: president jimmy carter turned all white house thermostats down to 65 degrees during the winter of 1977. the very first phone number of the white house was ``1''. jimmy carter had solar panels installed on the white house...and ronald reagan had them removed. there is a replica of the white house in atlanta which was built as a private home you can mail a birth announcement to the white house and they'll send you a congratulations card back.} & No

\\
\midrule
Groundedness & Dialogue &
\multicolumn{1}{m{9cm}}{\colorR{question}: Does this response use knowledge from the fact? </s> \colorG{response}: batman's city of gotham is located in new jersey and neither batman nor the villain the joker refer to one another by name. </s> \colorB{fact}: there was a batman villain named condiment king and he was defeated by slipping on his own ketchup. adam west has a batman logo on one of his molars according to dc canon; batman's gotham city is located in new jersey in their face to face confrontations, neither batman nor joker refer to one another by name. weird al yankovich did voiceover work in the most recent dc animated film ``batman vs robin''.} & Yes
\\

\bottomrule
\end{tabular}
\caption{Examples of different dimensions in evaluation tasks. The red text indicates the question $q$, the green text denotes the candidate output $x$, the yellow text is the reference text $y$, and the blue text represents the context $c$.}
\label{tab:example_eval}
\end{table*}

We provide the concrete examples for different dimensions of evaluation tasks in Table~\ref{tab:example_eval}. All the pseudo data is constructed on the CNN/DailyMail~\cite{hermann2015teaching} and Topical-Chat~\cite{gopalakrishnan2019topical} corpus.
We input reference text $y$ (green text) to the model only when evaluating the \texttt{relevance} dimension in text summarization, while in the other dimensions \textsc{UniEval} is a reference-free evaluator.
Depending on the specific dimension, we feed the model with different contexts $c$.
In addition, We use ``\textbackslash n'' to separate the different turns in the dialogue history and end it with ``\textbackslash n\textbackslash n''.

\section{Examples for Intermediate Tasks}
\label{sec:example_intermediate}
\renewcommand\arraystretch{1.0}
\begin{table*}
\centering \footnotesize
\label{tab:pattern}
\begin{tabular}{lccc}
\toprule
\textbf{Tasks} & \textbf{Datasets} & \textbf{Input} & \textbf{Target} \\
\midrule
NLI & DocNLI & \multicolumn{1}{m{9cm}}{\colorR{question}: Is this a claim consistent with the premise? </s> \colorB{claim}: The two victims were teenagers. </s> \colorB{premise}: 2 seriously wounded in Grand Crossing shooting Two men were seriously wounded in a shooting Thursday evening in the Grand Crossing neighborhood on the South Side. The men \u2014 ages 28 and 39 \u2014 were shot at by someone inside a vehicle that pulled up to them at 6:11 p.m. in the 7300 block of South Dante, Chicago Police said. The older man was shot in his face and taken to Northwestern Memorial Hospital in serious condition, police said. The younger man took himself to Jackson Park Hospital with a gunshot wound to his shoulder in serious condition.} & No \\
\midrule

NLI & MRPC & \multicolumn{1}{m{9cm}}{\colorR{question}: Is this sentence equivalent to the reference? </s> \colorB{sentence}: The DVD-CCA then appealed to the state Supreme Court. </s> \colorB{reference}: The DVD CCA appealed that decision to the U.S. Supreme Court.} & Yes \\
\midrule

NLI & QQP &
\multicolumn{1}{m{9cm}}{\colorR{question}: Is the following question equivalent to the reference? </s> \colorB{question}: Do you need a passport to go to Jamaica from the United States? </s> \colorB{reference}: How can I move to Jamaica?} & No \\
\midrule

SST & CNN/DailMail & \multicolumn{1}{m{9cm}}{\colorR{question}: Is this sentence the coherent first sentence of the document? </s> \colorB{sentence}: Diego Maradona thinks Steven Gerrard and England's defence should be held responsible for the defeat to Uruguay that crushed their World Cup hopes. </s> \colorB{document}: Bomb disposal experts examined the mortars and confirmed that two contained white phosphorous, a police spokesman said. It was not the first time that Hamas has attempted to target Israel using mortars containing white phosphorous, the spokesman said. White phosphorus ignites and burns, creating white smoke when it is exposed to oxygen. Militaries use it as a smoke screen to protect troops during combat ...} & No \\
\midrule

Linguistics & CoLA &
\multicolumn{1}{m{9cm}}{\colorR{question}: Is this a fluent and linguistically acceptable sentence? </s> \colorB{sentence}: Mary questioned Joe's desire to eat cabbage, but only after I had questioned Sally's desire to.} & No \\
\midrule

QA & BoolQ &
\multicolumn{1}{m{9cm}}{\colorR{question}: is confectionary sugar the same as powdered sugar? </s> \colorB{context}: Powdered sugar, also called confectioners' sugar, icing sugar, and icing cake, is a finely ground sugar produced by milling granulated sugar into a powdered state. It usually contains a small amount of anti-caking agent to prevent clumping and improve flow. Although most often produced in a factory, powdered sugar can also be made by processing ordinary granulated sugar in a coffee grinder, or by crushing it by hand in a mortar and pestle.} & Yes
\\
\midrule

QA & StrategyQA &
\multicolumn{1}{m{9cm}}{\colorR{question}: Is a Boeing 737 cost covered by Wonder Woman (2017 film) box office receipts? </s> \colorB{term}: Wonder Woman (2017 film) </s> \colorB{description of term}: American superhero film directed by Patty Jenkins </s> \colorB{facts}: The average cost of a US Boeing 737 plane is 1.6 million dollars. Wonder Woman (2017 film) grossed over 800 million dollars at the box office.} & Yes
\\
\midrule

QA & MultiRC & 
\multicolumn{1}{m{9cm}}{\colorR{question}: Did Susan's sick friend recover? </s> \colorB{context}: Sent 1: Susan wanted to have a birthday party. Sent 2: She called all of her friends. Sent 3: She has five friends. Sent 4: Her mom said that Susan can invite them all to the party. Sent 5: Her first friend could not go to the party because she was sick. Sent 6: Her second friend was going out of town. Sent 7: Her third friend was not so sure if her parents would let her. Sent 8: The fourth friend said maybe. Sent 9: The fifth friend could go to the party for sure. Sent 10: Susan was a little sad. Sent 11: On the day of the party, all five friends showed up. Sent 12: Each friend had a present for Susan. Sent 13: Susan was happy and sent each friend a thank you card the next week.} & Yes \\

\bottomrule
\end{tabular}
\caption{Examples for different intermediate tasks. Since these tasks are not relevant to the evaluation, we recognize all parts except question $q$ as context $c$.}
\label{tab:example_inter}
\end{table*}

We also include the examples for each intermediate task in Table~\ref{tab:example_inter}.
We define the input as a $(c, q)$ pair and let the model answer with ``Yes'' or ``No''.

\section{Implementation Details}
\label{sec:implementation}

\renewcommand\arraystretch{1.3}
\begin{table*}[t]
\center \footnotesize
\tabcolsep0.1 in
\begin{tabular}{lccccccccc}
\toprule
\multicolumn{1}{l}{\multirow{2}[1]{*}{\textbf{Metrics}}} & \multicolumn{3}{c}{\textbf{QAGS-CNN}}
 & \multicolumn{3}{c}{\textbf{QAGS-XSUM}} & \multicolumn{3}{c}{\textbf{Average}}  \\
 
 & $r$ & $\rho$ & $\tau$ &  $r$ & $\rho$ & $\tau$ & $r$ & $\rho$ & $\tau$ \\
 
\cmidrule(lr){1-1} \cmidrule(lr){2-4} \cmidrule(lr){5-7} \cmidrule(lr){8-10}

\textsc{ROUGE-1} & 0.338 & 0.318 & 0.248 & -0.008 & -0.049 & -0.040 & \cellcolor[rgb]{ .851,  .851,  .851} 0.165 & \cellcolor[rgb]{ .851,  .851,  .851} 0.134 & \cellcolor[rgb]{ .851,  .851,  .851} 0.104  \\

\textsc{ROUGE-2} & 0.459 & 0.418 & 0.333 & 0.097 & 0.083 & 0.068 & \cellcolor[rgb]{ .851,  .851,  .851} 0.278 & \cellcolor[rgb]{ .851,  .851,  .851} 0.250 & \cellcolor[rgb]{ .851,  .851,  .851} 0.200  \\

\textsc{ROUGE-L} & 0.357 & 0.324 & 0.254 & 0.024 & -0.011 & -0.009 & \cellcolor[rgb]{ .851,  .851,  .851} 0.190 & \cellcolor[rgb]{ .851,  .851,  .851} 0.156 & \cellcolor[rgb]{ .851,  .851,  .851} 0.122  \\

\textsc{BertScore} & 0.576 & 0.505 & 0.399 & 0.024 & 0.008 & 0.006 & \cellcolor[rgb]{ .851,  .851,  .851} 0.300 & \cellcolor[rgb]{ .851,  .851,  .851} 0.256 & \cellcolor[rgb]{ .851,  .851,  .851} 0.202  \\

\textsc{MoverScore} & 0.414 & 0.347 & 0.271 & 0.054 & 0.044 & 0.036 & \cellcolor[rgb]{ .851,  .851,  .851} 0.234 & \cellcolor[rgb]{ .851,  .851,  .851} 0.195 & \cellcolor[rgb]{ .851,  .851,  .851} 0.153  \\

\textsc{FactCC} & 0.416 & 0.484 & 0.376 & 0.297 & 0.259 & 0.212 & \cellcolor[rgb]{ .851,  .851,  .851} 0.356 & \cellcolor[rgb]{ .851,  .851,  .851} 0.371 & \cellcolor[rgb]{ .851,  .851,  .851} 0.294  \\

\textsc{QAGS} & 0.545 & - & - & 0.175 & - & - & \cellcolor[rgb]{ .851,  .851,  .851} 0.375 & \cellcolor[rgb]{ .851,  .851,  .851} - & \cellcolor[rgb]{ .851,  .851,  .851} - \\

\textsc{BartScore} & \textbf{0.735} & \textbf{0.680} & \textbf{0.557} & 0.184 & 0.159 & 0.130 & \cellcolor[rgb]{ .851,  .851,  .851} 0.459 & \cellcolor[rgb]{ .851,  .851,  .851} 0.420 & \cellcolor[rgb]{ .851,  .851,  .851} 0.343  \\

CTC (Consistency) & 0.619 & 0.564 & 0.450 & 0.309 & 0.295 & 0.242 & \cellcolor[rgb]{ .851,  .851,  .851} 0.464 & \cellcolor[rgb]{ .851,  .851,  .851} 0.430 & \cellcolor[rgb]{ .851,  .851,  .851} 0.346  \\

\midrule

\textsc{UniEval} (Consistency) & 0.682 & 0.662 & 0.532 & \textbf{0.461} & \textbf{0.488} & \textbf{0.399} & \cellcolor[rgb]{ .851,  .851,  .851} \textbf{0.571} & \cellcolor[rgb]{ .851,  .851,  .851} \textbf{0.575} & \cellcolor[rgb]{ .851,  .851,  .851} \textbf{0.465}  \\

\bottomrule
\end{tabular}
\caption{Pearson ($r$), Spearman ($\rho$) and Kendall-Tau ($\tau$) correlations of different metrics on QAGS benchmark.}
\label{tab:qags}
\end{table*}

We first train T5 on intermediate tasks for 2 epochs.
For the evaluation tasks, we construct pseudo data on the CNN/DailyMail~\cite{hermann2015teaching} and Topical-Chat corpus~\cite{gopalakrishnan2019topical} for summarization and dialogue generation, respectively.
The number of samples for each dimension is 30k, with an equal number of positive and negative examples.
We set batch size to 36 and the maximum learning rate to 5e-5 for both stages.
Regarding continuous learning, we randomly select 20\% of the data from the previously learned tasks to replay.
The order is \texttt{coherence} $\rightarrow$ \texttt{fluency} $\rightarrow$ \texttt{consistency} $\rightarrow$ \texttt{relevance} for summarization, and \texttt{coherence} $\rightarrow$ \texttt{naturalness} $\rightarrow$ \texttt{groundedness} $\rightarrow$ \texttt{engagingness} for dialogue generation.
Considering the difference in learning difficulty, we train 0.2-2.0 epochs for each dimension.
And for multi-task learning in the evaluation tasks, we train the evaluator for 1-3 epochs in different NLG tasks.
We train \textsc{UniEval} on two A6000 GPUs for a total of 5 hours.
If the meta-evaluation benchmark contains multiple references, we only use the first one as input.

In addition, although we can compute the scores for all dimensions directly from Equation \ref{equ:score}, we slightly modify the score calculation for several certain dimensions due to their characteristics.
For example, for \texttt{fluency} and \texttt{consistency} in summarization, disfluency and inconsistency are usually detected using sentences as the basic unit~\cite{fabbri2021summeval, laban2021summac}, so we split the model output $x$ into several sentences and calculate the score $s_{ij}$ for $j$-th sentence as:

\begin{align}
    s_{ij} = \frac{P(\mathrm{``Yes"}|x_j, y, c, q_i)}{P(\mathrm{``Yes"}|x_j, y, c, q_i)+P(\mathrm{``No"}|x_j, y, c, q_i)}.
\end{align}

Then the final score for $x$ in these two dimensions is $s_i = \sum_{j=1}^{m}s_{ij} / m$, where $m$ is the number of sentences in $x$.
Another special dimension is \texttt{engagingness} in dialogue generation.
Since it indicates the total volume of interesting facts presented in the response~\cite{deng2021compression}, we use the summation to compute it as $s_i = \sum_{j=1}^{m}s_{ij}$.
Therefore, the scoring range for \texttt{engagingness} is [0, $+\infty$), while all others are [0, 1].

\section{Results on QAGS}
\label{sec:qags}

Advanced NLG models suffer from the problem of generating text that is inconsistent with the source document~\cite{cao2018faithful}, which has led recent research to develop evaluators for evaluating the \texttt{consistency} dimension in summarization~\cite{kryscinski2020evaluating, wang2020asking, cao2020factual, durmus2020feqa}.
Therefore, we particularly compare the single-dimensional version of \textsc{UniEval} for \texttt{consistency} with the state-of-the-art factuality checkers.

We conduct experiments on the QAGS meta-evaluation benchmark, which contains two different summarizaion corpus: CNN/DailyMail~\cite{hermann2015teaching} and XSum~\cite{narayan2018don}.
As shown in Table~\ref{tab:qags}, BARTScore performs best on the more extractive\footnote{The reference summaries of CNN/DailyMail datasets tend to be copied from the original text.} part (QAGS-CNN), but shows poor correlation on the more abstractive\footnote{The words in the summary in Xsum dataset often do not appear in the original text.} subset (QAGS-Xsum).
\textsc{UniEval} (Consistency) correlates well in both parts of the data, especially in the more challenging Xsum dataset, greatly outperforming all previous consistency detectors.
On average, \textsc{UniEval} (Consistency) outperforms the state-of-the-art evaluator CTC by more than 30\% based on Spearman and Kendall-Tau correlations.
Thus, a high-performance single-dimensional evaluators can also be developed under our proposed framework.

\end{document}